\documentclass[10pt,twocolumn,letterpaper]{article}

\usepackage{cvpr}              
\usepackage{float}
\usepackage{algorithm}
\usepackage{algorithmic}
\usepackage{adjustbox}
\definecolor{cvprblue}{rgb}{0.21,0.49,0.74}
\usepackage[pagebackref,breaklinks,colorlinks,allcolors=cvprblue]{hyperref}

\def\paperID{*****} 
\def\confName{CVPR}
\def\confYear{2025}

\title{Efficient Self-Improvement in Multimodal Large Language Models: \\A Model-Level Judge-Free Approach}

\author{
Shijian Deng$^1$, Wentian Zhao$^2$, Yu-Jhe Li$^2$, Kun Wan$^2$, Daniel Miranda$^2$, Ajinkya Kale$^2$, Yapeng Tian$^1$ \\
$^1$The University of Texas at Dallas, $^2$Adobe Inc. \\
{\tt\small \{shijian.deng, yapeng.tian\}@utdallas.edu, \{wezhao, jhel, kuwan, miranda, akale\}@adobe.com}
}

\begin{document}
\maketitle
\begin{abstract}
Self-improvement in multimodal large language models (MLLMs) is crucial for enhancing their reliability and robustness. However, current methods often rely heavily on MLLMs themselves as judges, leading to high computational costs and potential pitfalls like reward hacking and model collapse.
This paper introduces a novel, model-level judge-free self-improvement framework. Our approach employs a controlled feedback mechanism while eliminating the need for MLLMs in the verification loop. We generate preference learning pairs using a controllable hallucination mechanism and optimize data quality by leveraging lightweight, contrastive language-image encoders to evaluate and reverse pairs when necessary.
Evaluations across public benchmarks and our newly introduced IC dataset—designed to challenge hallucination control—demonstrate that our model outperforms conventional techniques. We achieve superior precision and recall with significantly lower computational demands. This method offers an efficient pathway to scalable self-improvement in MLLMs, balancing performance gains with reduced resource requirements.
\end{abstract}    
\section{Introduction}
\label{sec:intro}

\begin{figure}[ht]
  \centering
  \includegraphics[width=1\linewidth]{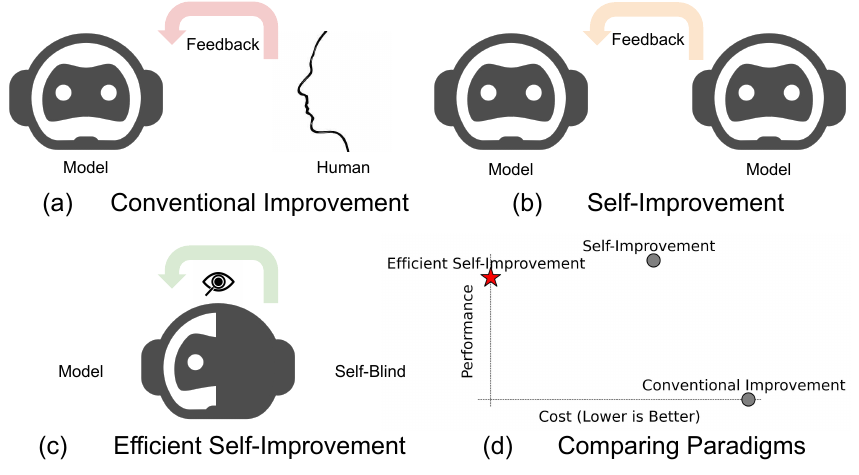}

  \caption{Comparison of three different improvement paradigms. (a) The conventional improvement paradigm requires humans to annotate feedback data and feed it into the model for improvement, making it the least efficient approach. (b) The self-improvement paradigm leverages the model itself to provide feedback; however, this approach is still inefficient due to the high cost and potential bias of using large models as verifiers. (c) Our efficient self-improvement paradigm improves the model without human feedback or model-level self-feedback by using a predefined data generation strategy combined with a lightweight verifier, achieving both efficiency and performance improvement. (d) Among all three paradigms, efficient self-improvement offers the best trade-off between performance and cost.}
  \label{fig:teaser}
\end{figure}

Self-improvement is a natural way for humans to learn independently, enabling them to acquire knowledge and skills beyond what they learn from their teachers. This same paradigm is being gradually adapted for large language models (LLMs) and multi-modal large language models (MLLMs) to achieve performance improvements beyond the seed model with minimal human supervision.

Recent studies have explored various approaches~\cite{favero2024multi, zhou2024aligning, deng2024enhancing, yu2024rlaif} to self-improvement in MLLMs. 
For instance, RLAIF-V~\cite{yu2024rlaif} uses MLLMs to evaluate and score responses generated by another MLLM, creating preference learning pairs from responses to the same image and question. M3ID~\cite{favero2024multi}, POVID~\cite{zhou2024aligning}, and STIC~\cite{deng2024enhancing} employ techniques like bad prompts, image corruption, unconditioned generation, and response injection to generate hallucinated responses as negative samples for preference learning.

However, several issues limit this paradigm: 1) it relies heavily on the quality of the verifier (\eg, a reward model); 2) the process can be resource-intensive, generating numerous samples but only using a tiny subset; 3) the cost multiplies when another large model is needed for verification, especially when generating reasoning or comments for final evaluation. Past studies~\cite{valmeekam2024planbench, valmeekam2024llms} have underscored the necessity of an external verifier.

To overcome these challenges, we propose an alternative approach, illustrated in Fig~\ref{fig:teaser}, enabling self-improvement without directly using an MLLM as a verifier for dataset filtering. Our method involves controlled hallucination to generate preference-learning pairs, lightweight evaluation with a contrastive language-image encoder to optimize data quality, and direct preference optimization (DPO)~\cite{rafailov2024direct} to train the seed model. 

First, we use an efficient, controllable approach to generate simple negative or hard-negative samples, creating the initial preference-learning pairs. We employ a controller ranging from 0 to 1 to control the level of hallucination in responses. After generating the initial dataset, we leverage a lightweight, contrastive language-image pre-trained encoder to compute average sentence-level CLIPScore~\cite{hessel2021clipscore}. This score identifies and updates pairs where the negative sample scores higher than the positive, refining our preference-learning dataset.
Finally, we use the optimized dataset to train the seed model via DPO~\cite{rafailov2024direct}, producing a self-improved model. Extensive evaluations on both in-house and public benchmarks show significant gains over the original seed model.

Our primary contributions are as follows:
\begin{itemize}
\item We propose a novel and efficient framework for self-improvement in MLLMs that: (a) combines a predefined, controllable mechanism for efficient negative sample generation, and (b) uses a lightweight verifier to effectively control positive and negative pairs, automatically reversing them when necessary.

\item We collected a new IC dataset, which includes GPT-4o-assisted evaluation both precision and recall of MLLMs.

\item Experimental results demonstrate that we can significantly better performance over the seed model on both our IC and Object HalBench datasets. 
\end{itemize}
\section{Related Work} 
\label{sec:related_work}

\subsection{Multimodal Large Language Models} To leverage the knowledge and reasoning capabilities of LLMs in multimodal settings and address broad multimodal comprehension challenges, MLLMs have been developed. Significant work has been done in this field, such as LLaVA~\cite{liu2024visual}, which connects CLIP with the LLaMA model through an adapter; Qwen-VL~\cite{bai2023qwen}, which implements grounding and text-reading abilities by aligning image-caption-box tuples; CogVLM~\cite{wang2023cogvlm}, which uses a trainable visual expert module in the attention and FFN layers to enable deep fusion of vision-language features without sacrificing NLP task performance; InternVL~\cite{chen2024internvl}, which employs both contrastive and generative tasks to better align the large-scale vision foundation model with MLLM; Pixtral~\cite{agrawal2024pixtral}, which processes images through the vision encoder at their native resolution and aspect ratio, converting them into image tokens for each patch in the image, allowing it to handle any number of images of arbitrary sizes in its large context window; and LLaMA3.2 Vision, which incorporates visual-recognition capabilities into LLaMA 3~\cite{dubey2024llama} via a compositional approach to ensure that text-only task performance is not affected by the addition of visual-recognition capabilities.

\subsection{Self-Improvement} Even after large-scale pretraining, instruction tuning, and reinforcement learning from human feedback (RLHF)~\cite{ouyang2022training}, large models may still show vulnerabilities in various cases. Although new data can always be prepared to improve a specific missing capability of the model, this is not a sustainable long-term solution to fix all issues at once. To enhance a large model’s helpfulness and trustworthiness without exhausting human effort, a new self-improvement paradigm has been adopted, as systematically discussed in the survey~\cite{tao2024survey}. For MLLMs, this often involves two key steps: sampling and verification.

\vspace{2mm}
\noindent
\textbf{Sampling.} To improve the seed model’s performance, the first step is to sample the necessary data. The simplest approach is to change seeds and randomly sample a large number of outputs, though this may not be efficient. Instead, users can predefine the type of data to generate by employing improved prompts and chains of thought to produce high-quality data, or by using corrupted images, attention masks, and text to generate negative data, as explored in POVID~\cite{zhou2024aligning}, STIC~\cite{deng2024enhancing}, and BDHS~\cite{amirloo2024understanding}. In M3ID~\cite{favero2024multi}, the authors also use mutual information from information theory to better control the quality of generated outputs and therefore achieve more effective sampling. In our work, we further simplify this sampling approach, making the process even more straightforward and practical.

\vspace{2mm}
\noindent
\textbf{Verification.} The model would not significantly improve if it simply reuses any generated data for retraining. A more effective approach is to perform data selection before training. There are many ways to achieve this. The simplest method is majority voting, though this may fail when the correct output is not the most common. A verifier, while optional, is commonly used as an additional quality control layer for data. The most straightforward and widely used verification method is to use an MLLM as a reward model, as seen in RLAIF-V~\cite{yu2024rlaif}. However, this approach has limitations related to cost and potential bias due to the reward models' own limitations. An external verifier can help address these issues. For example, CLIP-DPO~\cite{ouali2024clip} utilizes CLIP to rank short descriptions generated by the MLLM. We adopted a similar approach and extended it to suit long captions, seamlessly integrating it into our self-improvement framework along with our sampling methods to further enhance the robustness of our pipeline.

\section{Method}
\label{sec:method}

\begin{figure*}[thbp]
  \centering
   \includegraphics[width=0.9\linewidth]{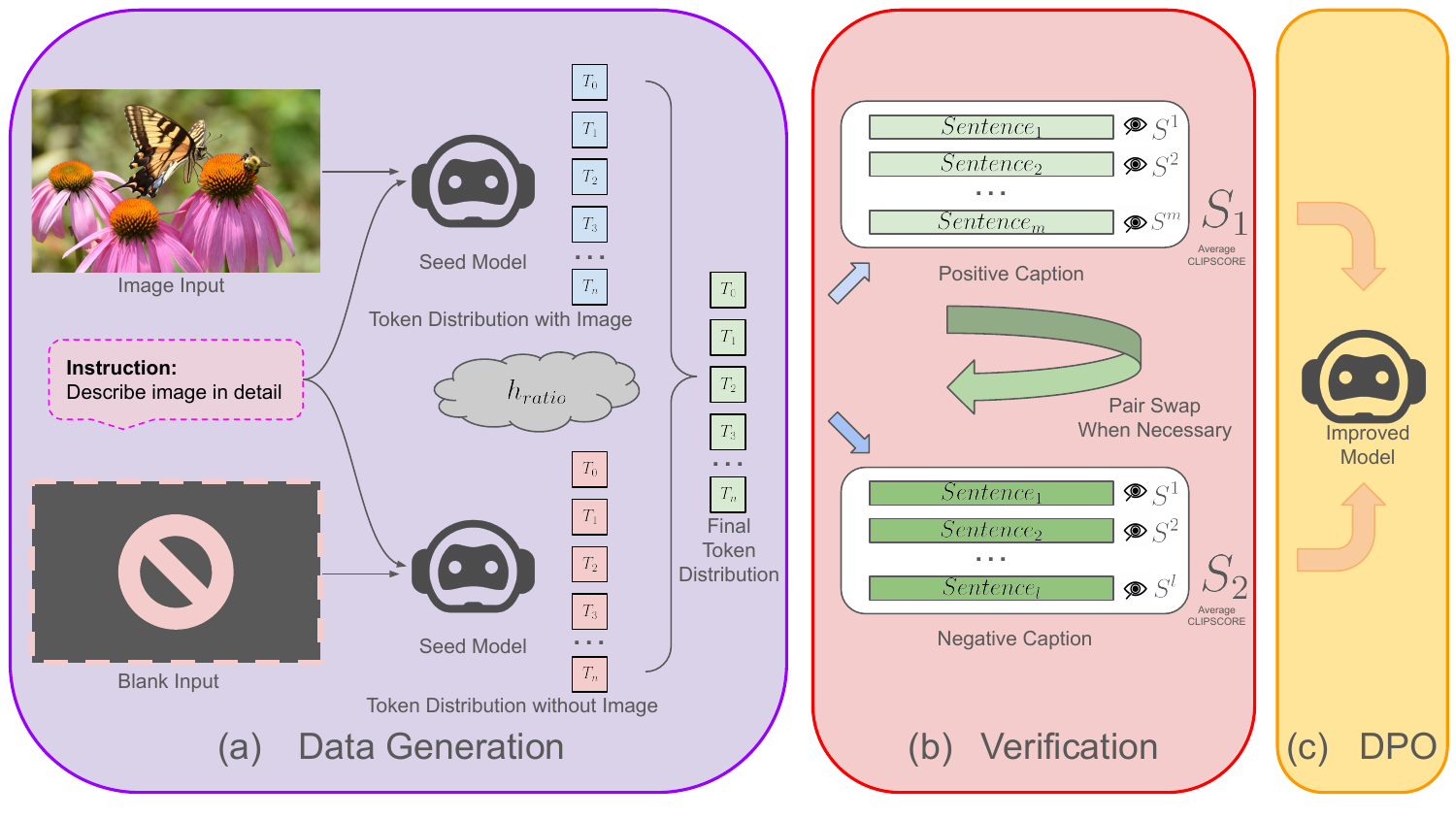}

    \caption{Overview of our framework. Our efficient self-improvement framework combines two main strategies: (a) We use a simple yet effective predefined preference dataset generation approach, employing two decoding paths during response generation. By adjusting the hallucination ratio $h_\text{ratio}$, we can control whether a negative or positive sample is generated for preference learning. (b) After the initial preferences are generated, we use a lightweight contrastive language-image pretrained encoder to calculate the average sentence-level $\text{CLIP}\_\text{score}$ difference between the initial positive and negative samples, swapping them when necessary to ensure the quality of the final preference dataset. (c) Finally, we apply DPO with the resulting dataset to improve the model.}
    
   \label{fig:method}
\end{figure*}

This section describes our approach to efficient self-improvement in MLLMs. We begin with a brief overview of DPO, followed by a description of our controllable method for generating positive and negative data pairs for training. Next, we highlight the importance of incorporating a lightweight quality control mechanism to ensure that the generated data effectively guides the learning process. Finally, we explain how the generated data is used to train the seed model with DPO, culminating in a self-improved model.

\subsection{Preliminaries: DPO}

DPO has recently emerged as a popular method for preference learning in large language models due to its simplicity compared to reinforcement learning-based techniques like RLHF and RLAIF. Unlike RLHF/RLAIF, which uses reinforcement learning to optimize a policy, DPO frames preference learning as a supervised learning task.

Given a dataset of preference pairs $(x, y_w, y_l)$, where $y_w$ is preferred over $y_l$ for input $x$, DPO directly optimizes model parameters $\theta$ to maximize the probability of preferred outputs and minimize the probability of dispreferred outputs relative to a reference model $\pi_0$.
The objective is often defined with a negative log-likelihood loss:

\begin{align}
    L(\theta) &= -\log \sigma\left( \Delta(x, y_w, y_l; \theta) \right),
\end{align}
where 
\begin{align}
\Delta(x, y_w, y_l; \theta) &= 
\left[ \log \pi_\theta(y_w \mid x) - \log \pi_\theta(y_l \mid x) \right] \nonumber \\
&\quad - \left[ \log \pi_0(y_w \mid x) - \log \pi_0(y_l \mid x) \right].
\end{align}
Here, $\pi_\theta$ represents the model parameterized by $\theta$, $\pi_0$ is the initial policy, $x$ is the input text prompt along with the image, and $\sigma$ is the sigmoid function.

This loss function encourages the model to assign higher probabilities to preferred responses and lower probabilities to dispreferred responses while staying close to the reference model. This approach avoids the complexity and instability of reinforcement learning, simplifying training and enhancing convergence.

In this work, we adopt DPO to optimize our model’s alignment with preference data efficiently generated and filtered by our framework for self-improvement.

\subsection{Motivation}

To perform well, preference learning requires diverse data and accurate preference labels for each pair, making it critical to establish a fully controllable approach for generating the required dataset. While it is challenging to produce data that surpasses the quality of what the seed model can generate, it is relatively feasible to create data that is worse than what the model can typically produce. It is also important to know how much worse the sample we need before we generate it since both too hard or too simple pairs may not work the best. Based on these observations, we propose a simple yet efficient method for generating preference learning data pairs with any difference level between positive samples and negative samples.

The high computational cost of running models with a large number of parameters, combined with the inherent inductive biases of MLLMs, imposes significant limitations on relying on large models as verifiers. To address these challenges, we introduce an objective and lightweight alternative for verification purposes.

\subsection{Controllable Dataset Generation}
To train a self-improving model with preference learning, we first need to prepare a suitable dataset. To generate preference pairs, we use the seed model $m_0$ to create a positive response $y_w$ and a negative response $y_l$ from the same input image $x_{img}$ and instruction $x_{instruct}$. To obtain the negative sample, we introduce interventions during the decoding process of the MLLM. We use two decoding paths: a conditional path $p_c$ that generates a response based on both the input image $x_{img}$ and instruction $x_{instruct}$, and an unconditional path $p_u$ that uses only the instruction $x_{instruct}$, without the image $x_{img}$. The generation is controlled by the hallucination ratio, \ie, $h_\text{ratio}$, which determines the level of hallucination to be injected into the generated caption, ranging from $0$ to $1$. A higher $h_\text{ratio}$ denotes injecting more hallucinations into the response.

As shown in Fig.~\ref{fig:method}, for each output token, the distribution is determined by combining the token distribution $t_c$ from the conditional path and the token distribution $t_u$ from the unconditional path, weighted by the hallucination ratio (\ie, $h_\text{ratio}$): 
\begin{align}
    t &= (1 - h_\text{ratio}) \cdot t_c + h_\text{ratio} \cdot t_u.
\end{align}
The two paths, $p_c$ and $p_u$, do not interact, ensuring that the unconditional path never accesses any information from the input image, thus serving as a "pure" hallucination source. Each pair is initially labeled, with the response generated under the lower $h_\text{ratio}$ assigned as positive and the other as negative. The $h_\text{ratio}$ follows a predefined distribution, such as uniform or Gaussian, and remains fixed for each decoding process once assigned.

\subsection{Lightweight Preference Data Inversion}

Although the generated pairs initially have assigned positive or negative labels, these labels may not always be accurate, as the conditional generation process with the seed MLLM can sometimes introduce a certain level of hallucination in the decoded text. To address this, we implement an additional quality control step to manage cases where initial labeling may be incorrect.

Specifically, we use a lightweight CLIP model, which is the vision-language contrastive pretrained encoder of the MLLM. For each initial pair $(y^i_w, y^i_l)$, we calculate the $\text{CLIP}\_\text{score}$ between the image and each decoded sentence. Since CLIP has a 77-token limit and cannot accommodate overly long captions, we compute the average sentence-level $\text{CLIP}\_\text{scores}$ for the initial positive caption, $\text{CLIP}\_\text{score}^i_w$, and the initial negative caption, $\text{CLIP}\_\text{score}^i_l$. If $\text{CLIP}\_\text{score}^i_w < \text{CLIP}\_\text{score}^i_l$, indicating that the initial positive is rated lower than the initial negative, we swap the preference labels, designating $y^i_w$ as the final negative $y^f_l$ and $y^i_l$ as the final positive $y^f_w$. Otherwise, we retain the original order in the final pair.

This process prevents cases where an initial negative sample might outperform its counterpart, which could undermine subsequent preference learning. After this step, we obtain the final preference pairs $(y^f_w, y^f_l)$, which are used in preference alignment training to improve the seed model $m_0$.

\subsection{Preference Learning Finetuning}
After obtaining the final pairs of positive caption $y^f_w$ and negative caption $y^f_l$ generated from the same input image $x_{img}$ and instruction $x_{instruct}$, we select a subset of the preference dataset $D$ within a certain range of the $\text{CLIP}\_\text{score}$ difference, $\text{CLIP}\_\text{score}^f_w - \text{CLIP}\_\text{score}^f_l$, forming $D_{sub}$. We then use DPO, a commonly used, low-cost alternative to RLHF, to train the seed model $m_0$, further enhancing its performance. 

Through this finetuning process, we obtain an improved model $m_1$, which is self-improved from the seed model $m_0$ using its own generated dataset. The detailed process is illustrated in Fig.~\ref{fig:method} and Algorithm~\ref{alg:self-improve}.

\begin{algorithm}[t]
\caption{Efficient Self-Improving MLLM with Preference Learning}
\label{alg:self-improve}
\begin{algorithmic}[1]
\REQUIRE Seed model $m_0$, dataset $\{(x_{\text{img}}^i, x_{\text{instruct}}^i)\}_{i=1}^N$
\ENSURE Improved model $m_1$

\FOR{each $(x_{\text{img}}, x_{\text{instruct}})$ in dataset}
    \STATE Sample hallucination ratio $h_\text{ratio} \in [0,1]$ from a predefined distribution
    \FOR{each time step $t$}
        \STATE Compute conditional token distribution $t_c = p_c(y_t \mid x_{\text{img}}, x_{\text{instruct}}, y_{<t})$
        \STATE Compute unconditional token distribution $t_u = p_u(y_t \mid x_{\text{instruct}}, y_{<t})$
        \STATE Compute final token distribution $t = (1 - h_\text{ratio}) \times t_c + h_\text{ratio} \times t_u$
        \STATE Sample token $y_t \sim t$
    \ENDFOR
    \STATE Obtain responses $y_{\text{low}}$ (lower $h_\text{ratio}$) and $y_{\text{high}}$ (higher $h_\text{ratio}$)
    \STATE Assign initial labels: positive response $y^i_w = y_{\text{low}}$, negative response $y^i_l = y_{\text{high}}$
    \STATE Compute average CLIP scores $CLIP\_score^i_w$ and $CLIP\_score^i_l$
    \IF{$CLIP\_score^i_w - CLIP\_score^i_l < 0$}
        \STATE Swap $y^i_w$ and $y^i_l$
    \ENDIF
    \STATE Add final pair $(x_{\text{img}}, x_{\text{instruct}}, y^f_w, y^f_l)$ to preference dataset $D$
\ENDFOR

\STATE Select subset $D_{\text{sub}}$ from $D$ based on $CLIP\_score$ difference
\STATE Initialize improved model $m_1 \leftarrow m_0$

\FOR{each $(x, y^w, y^l)$ in $D_{\text{sub}}$}
    \STATE Compute $\Delta(x, y^w, y^l; \theta) = [\log \pi_\theta(y^w \mid x) - \log \pi_\theta(y^l \mid x)] - [\log \pi_0(y^w \mid x) - \log \pi_0(y^l \mid x)]$
    \STATE Compute loss $L(\theta) = -\log \sigma\left( \Delta(x, y^w, y^l; \theta) \right)$
    \STATE Update model parameters $\theta$ by minimizing $L(\theta)$
\ENDFOR

\end{algorithmic}
\end{algorithm}

\section{Experiments}
\label{sec:experiments}

To evaluate the effectiveness of our proposed self-improvement framework, we tested it on both our IC dataset using GPT-4o series evaluation and a commonly used benchmark. We introduce the experimental settings for dataset generation and verification, followed by a detailed analysis of results and ablation studies to demonstrate the effectiveness of our framework and each of its design modules.

\subsection{Datasets}

\noindent
\textbf{IC Dataset.} Current hallucination benchmarks primarily evaluate the precision of captions while often ignoring recall. To comprehensively assess MLLMs’ captioning abilities, we have collected a new dataset containing 150 challenging images prone to hallucination across a wide range of domains and scenarios. These include abstract concepts, animals, animations, artistic content, common sense violations, documents, events, fashion, food, handwriting, illustrations, objects, people, posters, scenes, technology, and vehicles. 

After generating captions, we use the GPT-4o series to evaluate them based on precision (elements in the caption that are present in the image) and recall (elements in the image that are captured in the caption) to calculate a final F1 score, which serves as a measure of caption quality. Some examples are shown in Fig.~\ref{fig:qualitative_results}.

\vspace{2mm}
\noindent
\textbf{Object HalBench.}
In addition to our newly collected IC dataset, we selected a commonly used public benchmark to evaluate the improved model's performance: Object HalBench~\cite{rohrbach2018object}, a classic benchmark that focuses on evaluating object-level hallucination in vision-language models and is widely used to assess MLLM trustworthiness.

\subsection{Experiment Setup}
For the seed model $m_0$, we used LLaVA-1.5-13B~\cite{liu2024improved}, a popular and representative MLLM. An 8xA100 node with 80GB VRAM per GPU was used for DPO training, while data generation and other processes were performed on a single GPU.

During data generation, we sampled 100k images from the LLaVA instruction tuning dataset, llava\_v1\_5\_mix665k, and removed all question-answer pairs. Using the prompt "Describe image in detail," the model generated responses with an $h_\text{ratio}$ ranging from $0$ to $1$. Initially, captions generated with a lower $h_\text{ratio}$ were assigned as the initial positive samples, $y^i_w$, while captions generated from the same inputs $x_\text{instruct}$ and $x_\text{img}$ with a higher $h_\text{ratio}$ were assigned as the initial negative samples, $y^i_l$. This process resulted in 100k initial preference pairs.

For the obtained caption pairs, each sentence was extracted, and the CLIP model was used to compute the $\text{CLIP}\_\text{score}$ for each image-caption pair. For sentences longer than the CLIP model’s context limit, we split them into shorter sub-sentences, computed their respective $\text{CLIP}\_\text{scores}$, and calculated the average $\text{CLIP}\_\text{score}$ for each caption by averaging the scores from all sentences and sub-sentences. If the average $\text{CLIP}\_\text{score}$ of a negative caption was higher than that of the positive caption, we swapped the positive and negative samples. The pairs were then sorted by $\text{CLIP}\_\text{score}$ difference, from low to high, and organized into 10 splits, each containing 10k pairs.

For each split, we trained a LLaVA model and conducted inference on the IC dataset and other benchmarks to gather results. For the IC dataset, GPT-4o was used as the evaluator to compute precision, recall, and F1 score.

\begin{figure*}[ht]
  \centering
   \includegraphics[width=1\linewidth]{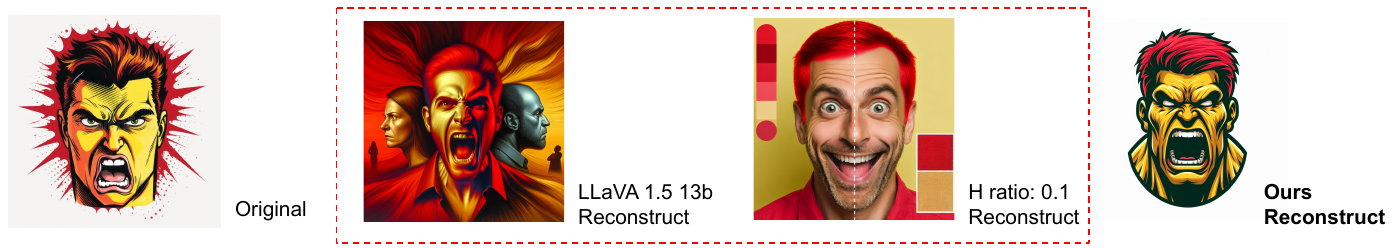}

   \caption{Image reconstruction examples. To further demonstrate the effectiveness of our training framework, we use a text-to-image diffusion model, DALL·E 3, to convert captions generated by the models back into images. The reconstructed image from the original model's caption contains significant hallucination, while increasing the $h_\text{ratio}$ during generation produces a negative caption that, when reconstructed, shows even more hallucination in attributes like style and emotion. However, after training with these generated caption pairs, the reconstructed image from the improved model’s caption closely resembles the original, surpassing both the positive and negative samples.}
   \label{fig:image_reconstruction}
\end{figure*}

\subsection{Results}

With the improved model $m_1$ derived from the original seed model $m_0$, we evaluated performance across different benchmarks and presented the results in Tab~\ref{tab:model_comparison_reduced} and Tab~\ref{tab:ic_dataset}. As shown, the self-improved model outperforms previous models on both benchmarks. In particular, compared to the original seed model LLaVA-1.5-13B, performance has improved significantly, clearly demonstrating the effectiveness of our framework. Compared to previous methods, ours is the first to emphasize an efficient self-improvement approach that balances efficiency and effectiveness.

\begin{table}[t]
\centering
\adjustbox{max width=\columnwidth}{
    \begin{tabular}{l|cccc}
\toprule
Model & Size & Feedback & \multicolumn{2}{c}{Object HallBench} \\
\cmidrule(lr){4-5}
 &  &  & Resp. ↓ & Ment. ↓ \\
\midrule
VCD~\cite{leng2024mitigating} & 7B & No & 48.8 & 24.3 \\
OPERA~\cite{huang2024opera} & 7B & No & 45.1 & 22.3 \\
Less-is-more~\cite{yue2024less} & 7B & No & 40.3 & 17.8 \\
LURE~\cite{zhou2023analyzing} & 7B & No & 27.7 & 17.3 \\
QWEN-VL~\cite{bai2023qwen} & 10B & No & 40.4 & 20.7 \\
MiniGemini~\cite{li2024mini} & 34B & No & 14.5 & 8.0 \\
LLaVA-NeXT~\cite{liu2024llava} & 34B & No & 12.6 & 6.4 \\
\midrule
HA-DPO~\cite{zhao2023beyond} & 7B & Rule & 39.9 & 19.9 \\
POVID~\cite{zhou2024aligning} & 7B & Rule & 48.1 & 24.4 \\
Silkie~\cite{li2023silkie} & 10B & GPT-4V & 27.1 & 13.4 \\
LLaVA-RLHF~\cite{sun2023aligning} & 13B & Human & 38.1 & 18.9 \\
RLHF-V~\cite{yu2024rlhf} & 13B & Human & 12.2 & 7.5 \\
\midrule
LLaVA 1.5~\cite{liu2024improved} & 7B & No & 53.6 & 25.2 \\
LLaVA 1.5~\cite{liu2024improved} & 13B & No & 51.6 & 24.6 \\
+ Ours & 13B & Self-Efficiency & \textbf{9.4} & \textbf{5.1} \\
\bottomrule
\end{tabular}
}
\caption{Main results of our experiments on Object HallBench. Comparison of various models across different metrics. Resp. indicates the response-level metric, and Ment. represents the mention-level metric. The best results are \textbf{highlighted}.}
\label{tab:model_comparison_reduced}
\end{table}

\begin{table}[t]
\small
\setlength{\tabcolsep}{5pt}
\centering
\begin{tabular}{l|cccc}
\toprule
Model & Precision ↑ & Recall ↑ & F1 ↑ & Change ↑ \\ 
\midrule
LLaVA 1.5 13B & 6.6 & 6.56 & 6.58 & 0.00 \\ 
+ Ours & \textbf{7.74} & \textbf{7.78} & \textbf{7.76} & \textbf{1.18} \\ 
\bottomrule
\end{tabular}
\caption{Main results on the IC dataset. \textbf{Precision} measures how many elements in the caption are also in the image (higher scores indicate lower hallucination in the caption). \textbf{Recall} measures how many elements in the image are included in the caption, providing a complementary metric for hallucination evaluation. \textbf{F1} is the harmonic mean of precision and recall. All scores are on a scale from 1 (worst) to 10 (best). The best scores are highlighted.}
\label{tab:ic_dataset}
\end{table}

\begin{figure}[t]
  \centering

   \includegraphics[width=1\linewidth]{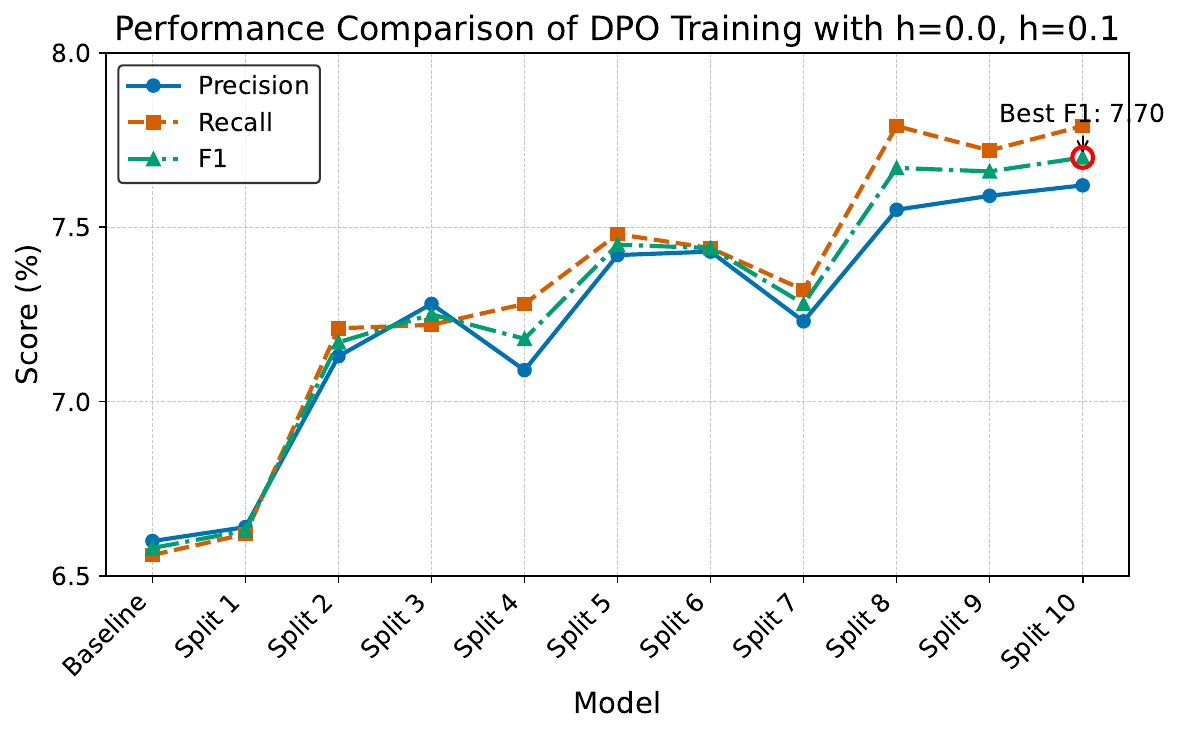}

   \caption{Performance comparison of DPO training using various $\text{CLIP}\_\text{score}$ differences generated with $h_\text{ratio}=0.0$ and $h_\text{ratio}=0.1$, ranked from low to high. The best performance is highlighted.}
   \label{fig:ablation_study}
\end{figure}

\begin{figure}[t]
  \centering

   \includegraphics[width=1\linewidth]{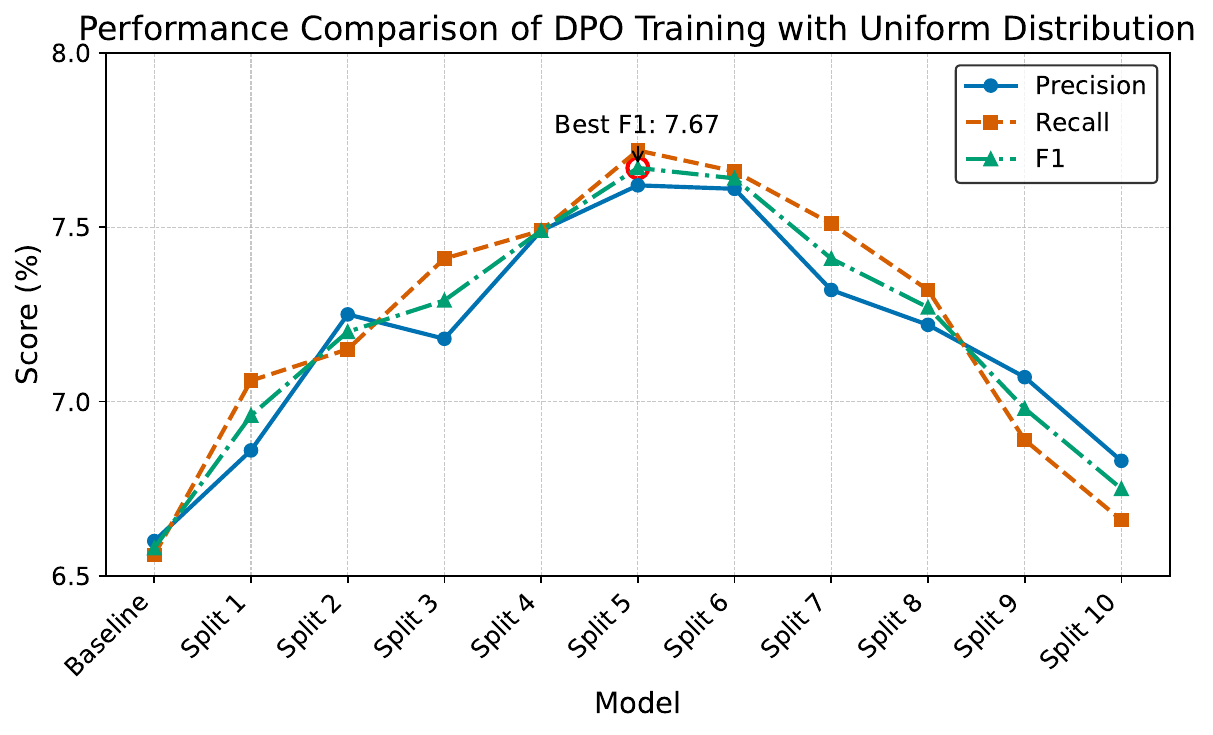}
    \vspace{-5mm}
   \caption{Performance comparison of DPO training using various $\text{CLIP}\_\text{score}$ differences generated with $h_\text{ratio}$ sampled from a uniform distribution and ranked from low to high. The best performance is highlighted.}
   \label{fig:ablation_study2}
\end{figure}

\begin{figure}[ht]
  \centering

   \includegraphics[width=1\linewidth]{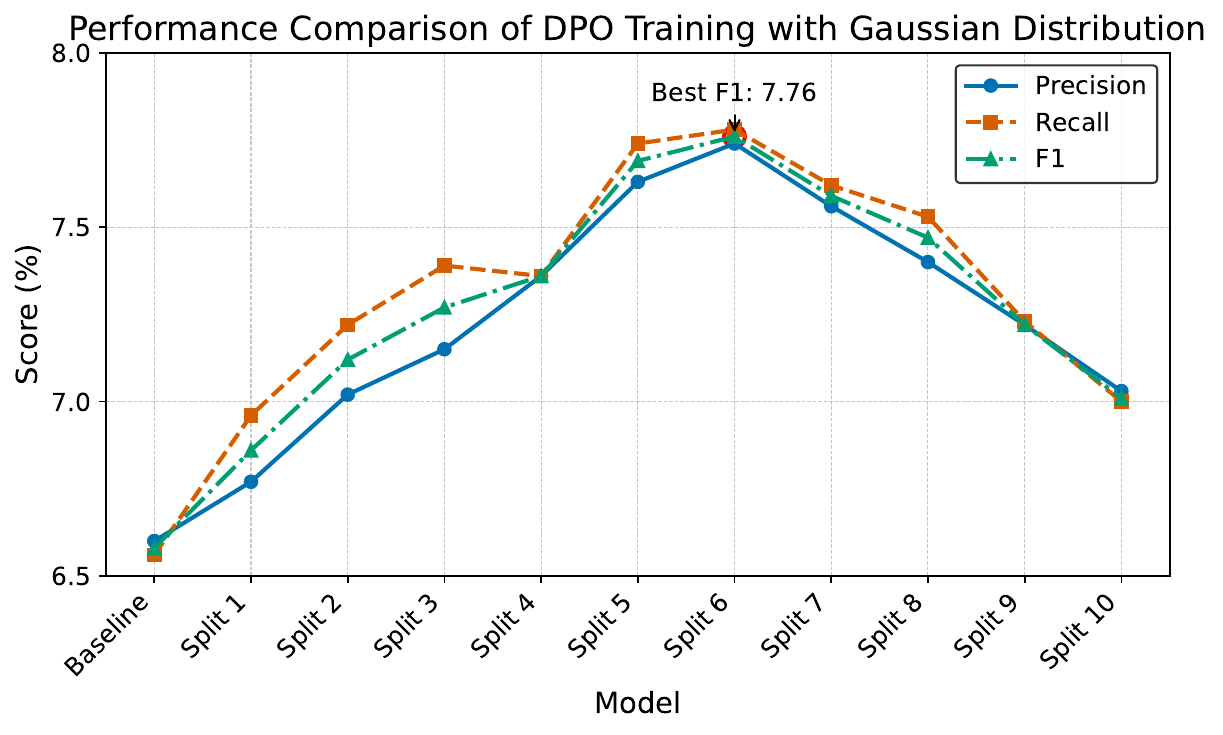}
    \vspace{-5mm}
   \caption{Performance comparison of DPO training using various $\text{CLIP}\_\text{score}$ differences generated with $h_\text{ratio}$ sampled from a Gaussian distribution and ranked from low to high. For these experiments, we set $\mu = 0.5$ and $\sigma = 0.15$. }
   \label{fig:ablation_study3}
\end{figure}

\begin{figure}[ht]
  \centering

   \includegraphics[width=1\linewidth]{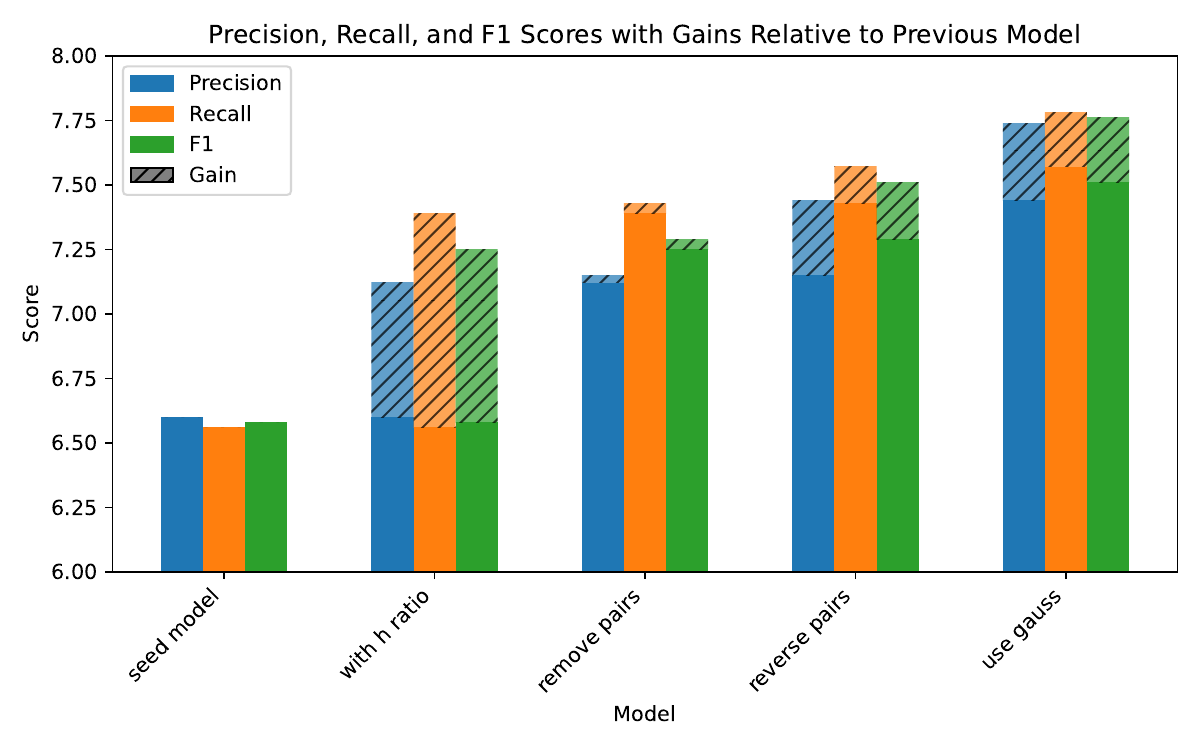}
\vspace{-7mm}
   \caption{Comparison of model performance when adding certain mechanisms in our framework. The striped sections indicate the performance gain after adding each module. We use LLaVA 1.5 13B as the seed model here. The modules gradually added to the seed model are: using $h_\text{ratio}$, removing pairs with negative $\text{CLIP}\_\text{score}$ differences, swapping pairs instead of removing them, and adding Gaussian distribution sampling. Each module in our design contributes to the final performance.}
   \label{fig:ablation_study4}
\end{figure}

For a fixed $h_\text{ratio}$ during dataset generation, we show the ablation study results in Fig.~\ref{fig:ablation_study}. Using a uniform distribution, we obtained the experimental results as illustrated in Fig.~\ref{fig:ablation_study2}. With different $\text{CLIP}\_\text{score}$ difference pairs generated with $h_\text{ratio}$ sampled from a Gaussian distribution, we also present the ablation study results in Fig.~\ref{fig:ablation_study3}. We observe a clear performance gain for each component added to our framework, compared to the seed model $m_0$ and the model without that component, as shown in Fig.~\ref{fig:ablation_study4}.

\begin{figure*}[htbp]
  \centering

   \includegraphics[width=0.95\linewidth]{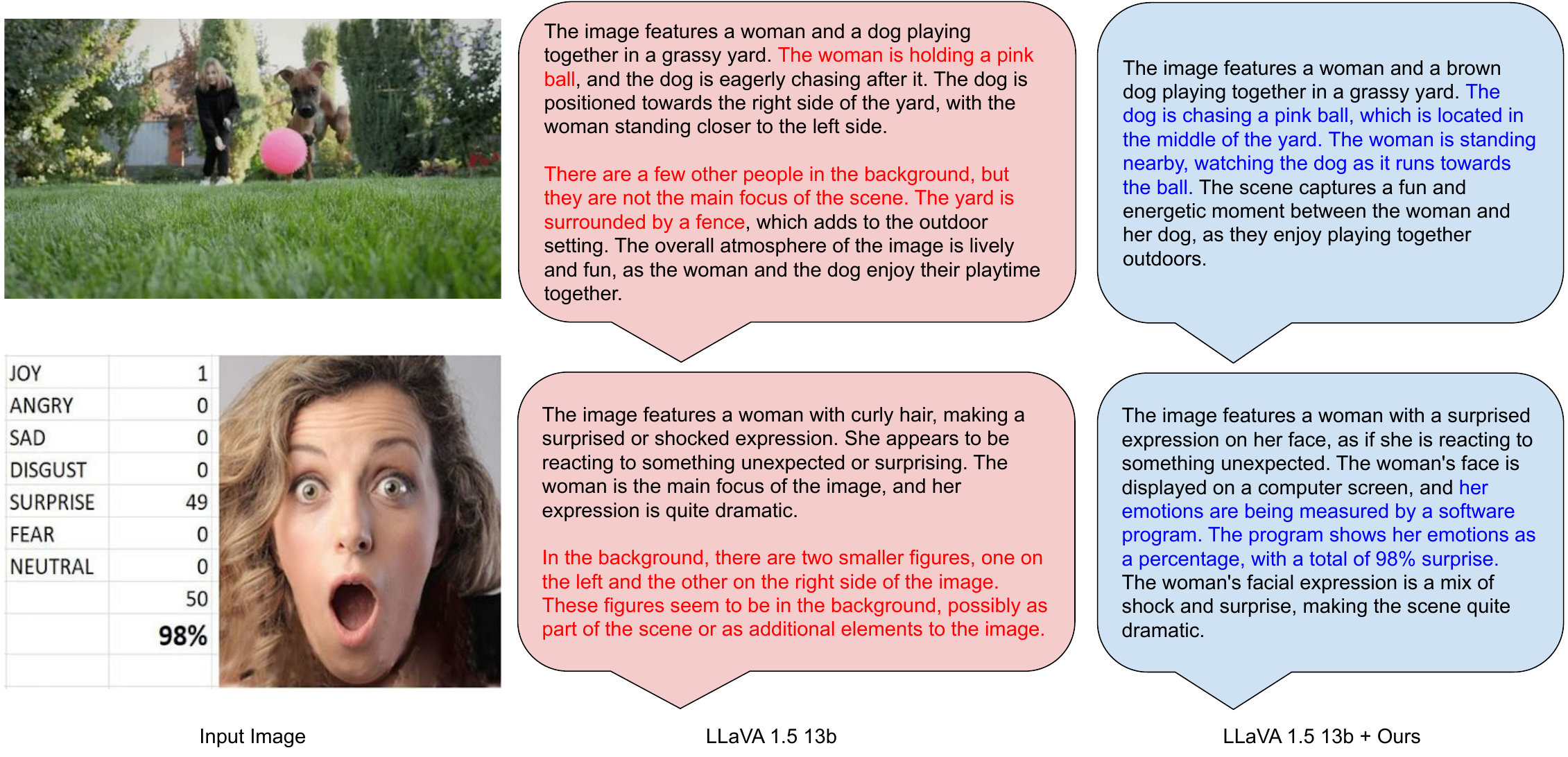}
    
   \caption{Examples of qualitative results. With the same input image and instruction prompt "Describe image in detail," the caption generated by the original seed model, LLaVA 1.5 13B, contains many hallucinations. In contrast, the model trained through our efficient self-improvement framework describes the image accurately, without hallucinated content. Hallucinated content is highlighted in red, and accurate content is highlighted in blue for easy identification.}
   \label{fig:qualitative_results}
   \vspace{-3mm}
\end{figure*}

In Fig.~\ref{fig:qualitative_results}, we show qualitative results to demonstrate the differences between our self-improved model and the original seed model. We also use the generated captions to perform image reconstruction with the DALL·E 3 model, as shown in Fig.~\ref{fig:image_reconstruction}.

\subsection{Experimental Analysis}

From the comprehensive evaluation results, we observe that our self-improved model shows significant performance gains across various benchmarks compared to the initial seed model in all evaluation dimensions. Here are some detailed discussions and findings from our experiments:

\vspace{2mm}
\noindent
 \textbf{Our model performs substantially better than the initial seed model,} as demonstrated by both quantitative and qualitative results. In Tab.~\ref{tab:model_comparison_reduced}, our model achieves scores of 9.4 for object response level and 5.1 for mention level, ranking it among the best of all models. These results on a popular public benchmark for evaluating hallucination demonstrate that our model outperforms all others at multiple evaluation levels, despite not using additional human feedback during finetuning or requiring feedback from an MLLM. Instead, it relies on lightweight CLIP encoders, highlighting the efficiency and effectiveness of our proposed framework.

\vspace{2mm}
\noindent
\textbf{Each of our designed modules contributes to performance improvement.} As shown in Fig.~\ref{fig:ablation_study4}, using $h_\text{ratio}$ to generate preference learning pairs and fine-tuning the seed model with DPO results in a substantial gain in both precision and recall compared to the original model. Adding $\text{CLIP}\_\text{score}$ difference filtering to exclude pairs with negative differences further enhances the model's performance. Instead of discarding these pairs, swapping the positive and negative samples when their $\text{CLIP}\_\text{score}$ difference is negative leads to another notable improvement. This highlights the necessity of a lightweight, post-hoc guard with $\text{CLIP}\_\text{score}$ difference to further boost performance. Finally, rather than using a fixed $h_\text{ratio}$, we experimented with a more diverse approach by randomly sampling $h_\text{ratio}$ values from a Gaussian distribution. This method introduces greater diversity, likely because it encompasses a broader range of cases, making the dataset more generalizable. This approach further improves the model’s performance by increasing the variety of negative samples.

\vspace{2mm}
\noindent
\textbf{The visual-language correspondence difference matters.} To evaluate model performance across settings with different preference pair combinations, we conducted extensive experiments varying the average sentence-level $\text{CLIP}\_\text{score}$ differences using fixed $h_\text{ratio}$ (see Fig.~\ref{fig:ablation_study}, which shows performance trends), uniform distribution sampling (see Fig.~\ref{fig:ablation_study2}, illustrating random variations), and Gaussian distribution sampling (see Fig.~\ref{fig:ablation_study3}, highlighting structured variability). The results indicate that selecting an optimal $\text{CLIP}\_\text{score}$ difference is critical, with performance peaking when differences are moderate—neither too large nor too small. This aligns with human learning patterns, where understanding improves most when examples are distinct enough to differentiate yet similar enough to allow meaningful comparisons. This insight opens an intriguing direction for future research: determining the optimal degree of difference between preference learning pairs to maximize learning efficiency.

\vspace{2mm}
\noindent
\textbf{Hallucinations reduction and better reconstructions.} From the qualitative examples in Fig.~\ref{fig:qualitative_results} and the reconstruction results in Fig.~\ref{fig:image_reconstruction}, we observe that while the seed model tends to hallucinate significantly, our model generates far more accurate content when provided with the same image and text prompt. These results also demonstrate how hallucinations can impair the reconstruction of an original image given a caption from a hallucination-prone model, and how our approach mitigates this issue. This could potentially contribute to building better reconstruction or generation models by using captions generated by our model.

\section{Limitations and Future Work}
\label{sec:limitations}

Although our experiments demonstrate that our framework is highly effective for enhancing the initial model’s performance, we acknowledge some limitations and highlight areas for exploration and improvement in future work.

\vspace{2mm}
\noindent
\textbf{Recursive Self-Improvement.} Due to limited resources, we were unable to investigate whether recursive self-improvement is feasible by iteratively applying our framework in multiple rounds, from data generation to preference learning finetuning, to go beyond $m_1$ and potentially achieve $m_2$, $m_3$, and so on. This could reveal whether further improvements are possible or if an upper performance bound exists.

\vspace{2mm}
\noindent
\textbf{Scaling with Larger Models and Datasets.} Because of training costs, we were unable to experiment with even larger models or larger datasets. Exploring the scaling laws of the framework with additional resources would be an interesting avenue for future research.

\vspace{2mm}
\noindent
\textbf{Extending to Other Modalities.} Although our experiments focused solely on vision-language tasks, the framework should be able to extend to other modalities, such as video and audio. These directions present promising topics for future exploration.

\section{Conclusion}
\label{sec:conclusion}

In this paper, we propose a novel and efficient self-improvement framework for MLLMs that does not require model-level self-feedback. We demonstrate that, using our methods: 1) We significantly improve the seed model's performance, reduce hallucination, and enhance image-caption correspondence compared to the original seed model across different benchmarks. 2) Our approach enables precise control over the pair generation process, allowing us to efficiently generate preference pairs with any desired level of difference between samples. 3) We prevent cases where the positive sample is worse than the negative one by using a lightweight CLIP model to flip samples when the score difference is negative. Unlike traditional self-improvement methods, our approach dramatically reduces the parameters required during the verification process, as it eliminates the need for a model-level judge. Extensive experiments demonstrate that our framework effectively balances superior performance and efficiency. We hope our work inspires new strategies for managing trade-offs in the self-improvement process for MLLMs.

{
    \small
    \bibliographystyle{ieeenat_fullname}
    \bibliography{main}
}

\clearpage
\setcounter{page}{1}
\maketitlesupplementary
\setcounter{section}{0}
\renewcommand{\thesection}{\Alph{section}}

In this appendix, we first provide additional details about our IC dataset, including image counts across its various categories. Next, we present additional qualitative results from our experiments, utilizing our efficient self-improvement framework in comparison to the seed model. Finally, with GPT-4o's evaluation, we further demonstrate the effectiveness of our proposed method.

\section{Details of the IC Dataset}

As mentioned in the main paper Sec~\ref{sec:experiments}, to comprehensively evaluate the model's performance across different caption cases, including the most challenging types, it was necessary to build a diverse dataset to address this issue.

We provide details of each category and the number of samples collected in our IC dataset in Table~\ref{tab:category_counts}.

\begin{table}[h!]
\centering
\begin{tabular}{|l|r|}
\hline
\textbf{Category} & \textbf{Count} \\ \hline
abstract          & 3  \\ \hline
animal            & 9  \\ \hline
animation         & 7  \\ \hline
artistic          & 7  \\ \hline
common            & 13 \\ \hline
documents         & 12 \\ \hline
events            & 10 \\ \hline
fashion           & 9  \\ \hline
food              & 9  \\ \hline
handwritten       & 5  \\ \hline
illustration      & 9  \\ \hline
object            & 12 \\ \hline
people            & 10 \\ \hline
poster            & 7  \\ \hline
scenes            & 9  \\ \hline
technology        & 9  \\ \hline
veichle           & 10 \\ \hline
\textbf{Total}    & \textbf{150} \\ \hline
\end{tabular}
\caption{Category and image counts of our IC dataset.}
\label{tab:category_counts}
\end{table}

\begin{figure*}[ht]
  \centering
   \includegraphics[width=1\linewidth]{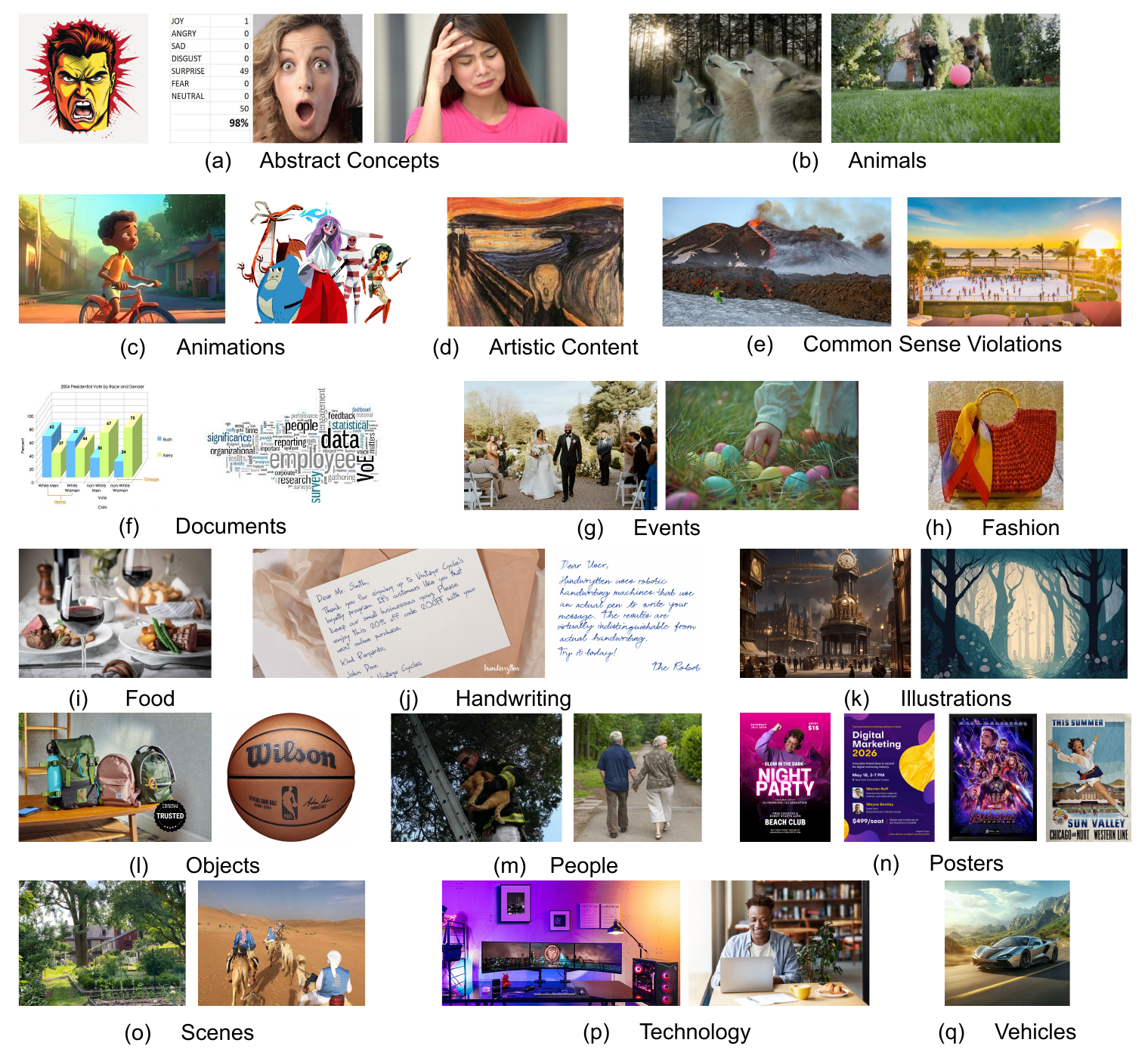}

    \caption{Examples from the IC dataset. The dataset covers a wide range of diverse categories important for image captioning tasks, including abstract concepts, animals, animations, artistic content, common sense violations, documents, events, fashion, food, handwriting, illustrations, objects, people, posters, scenes, technology, and vehicles.}
   \label{fig:ic_dataset}
\end{figure*}

More example visualizations of our proposed IC dataset can be found in Fig.~\ref{fig:ic_dataset}.

\section{Demo Examples}

\begin{figure*}[ht]
  \centering
   \includegraphics[width=1\linewidth]{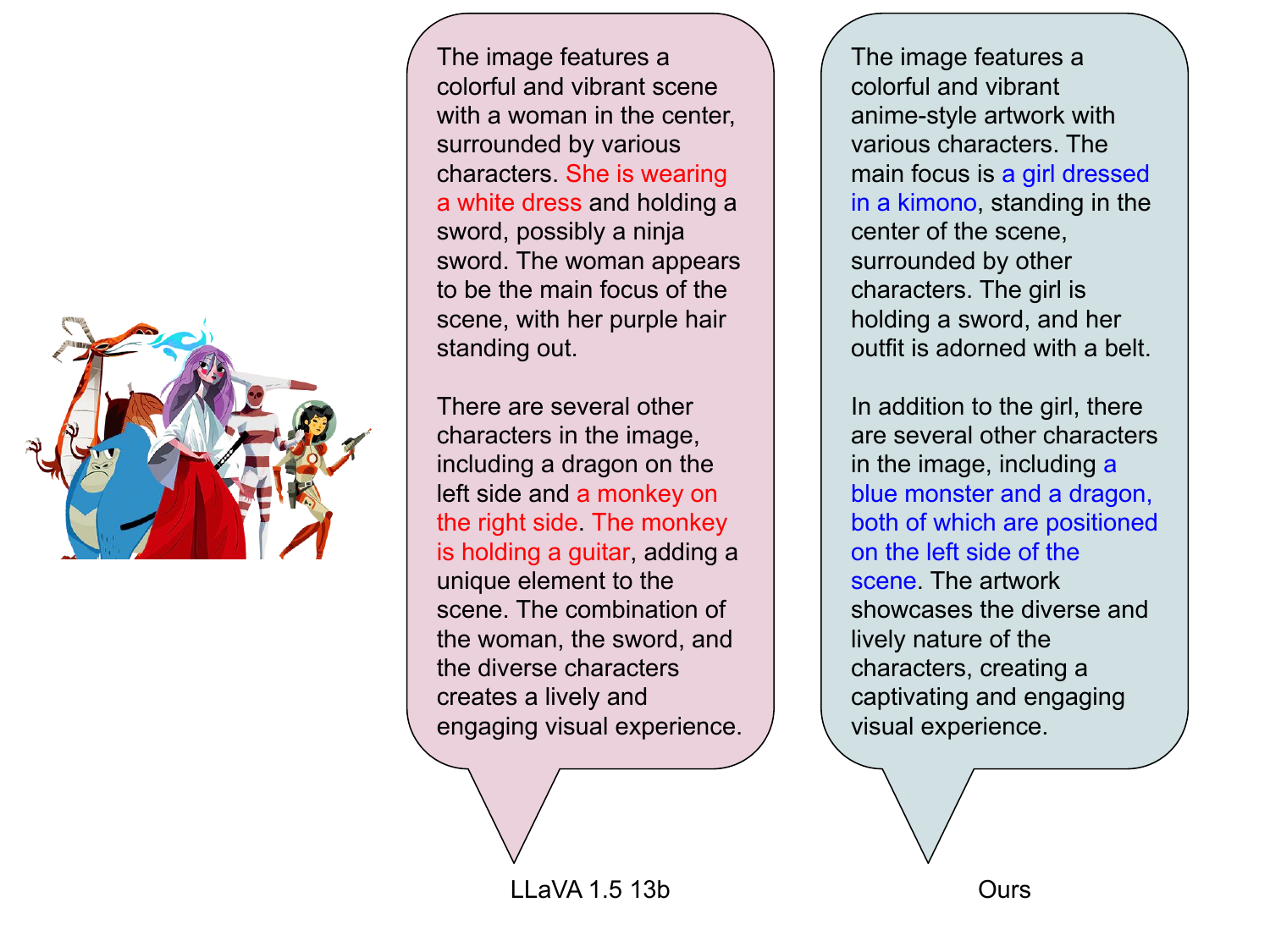}

   \caption{Qualitative comparison of animation image captions. The left panel shows the description generated by LLaVA 1.5 13b, while the right panel presents our method's output. Both descriptions capture the vibrant, anime-style scene featuring diverse characters, but our method demonstrates superior accuracy and specificity in detailing the central figure (a girl in a kimono) and surrounding elements (a blue monster and a dragon on the left). This highlights our approach's improved understanding of nuanced visual elements and context, enabling richer and more precise visual descriptions.}
   \label{fig:qualitative_results_1}
\end{figure*}

\begin{figure*}[ht]
  \centering
   \includegraphics[width=1\linewidth]{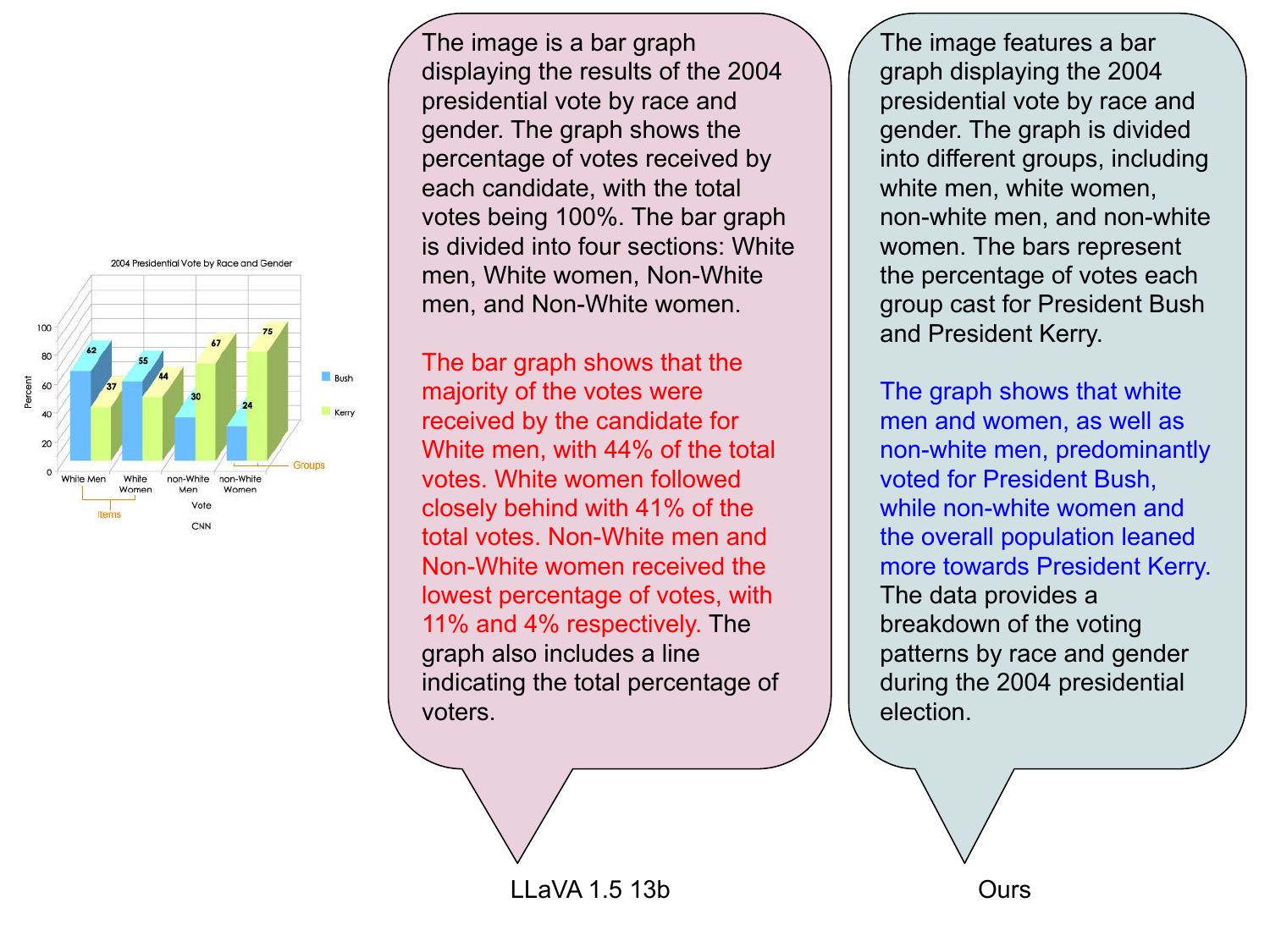}

   \caption{Qualitative comparison of document image captions. The left panel contains the description generated by LLaVA 1.5 13b, while the right panel shows the output from our method. Both descriptions analyze a bar graph depicting the 2004 U.S. presidential vote by race and gender, but our method provides a more concise and focused interpretation. It highlights key trends such as voting patterns for Bush and Kerry across demographic groups, emphasizing accuracy and clarity in extracting meaningful insights from visualized data.}
   \label{fig:qualitative_results_2}
\end{figure*}

\begin{figure*}[ht]
  \centering
   \includegraphics[width=1\linewidth]{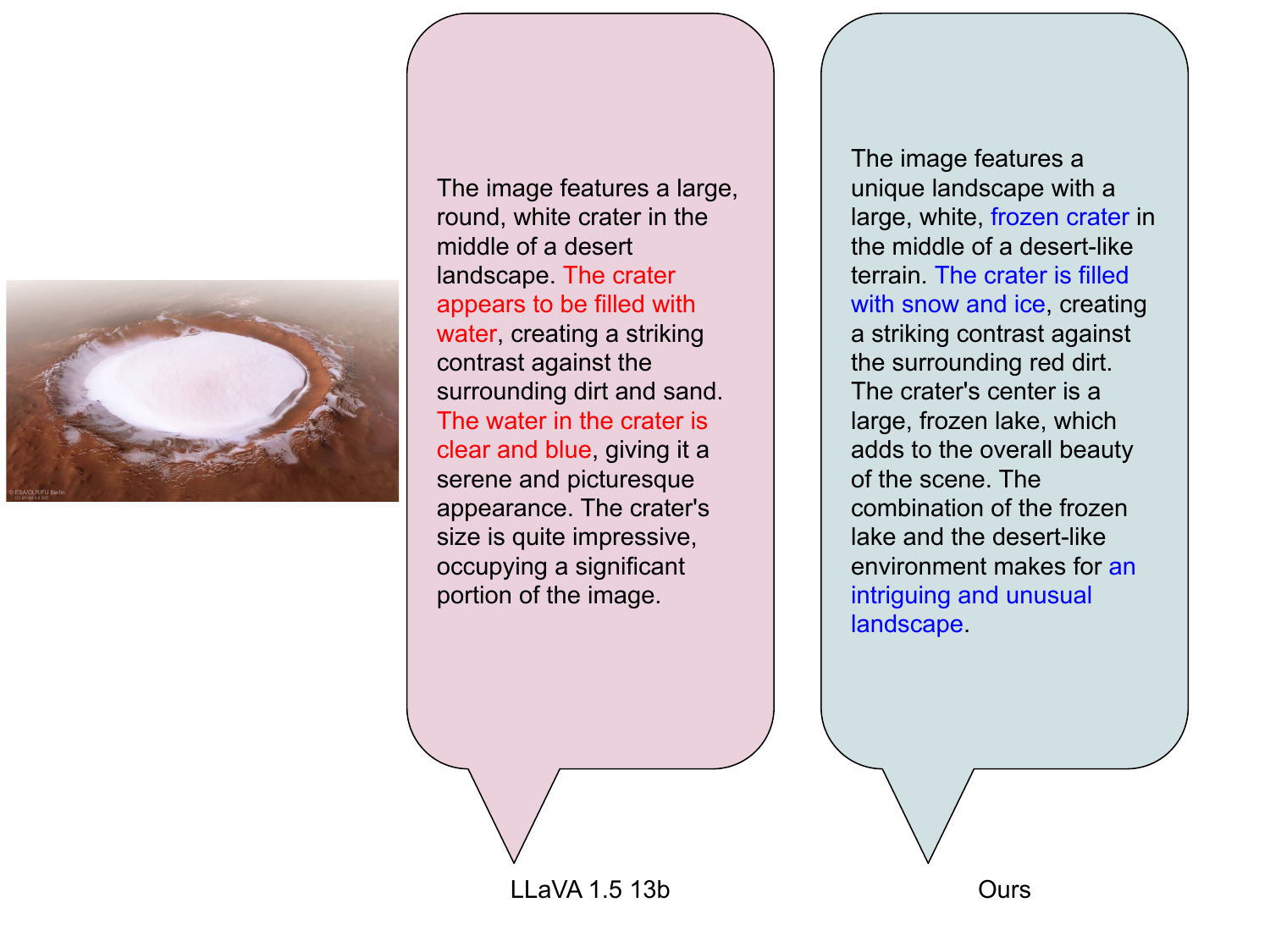}

   \caption{Qualitative comparison of common sense violation image captions. The left panel shows the description generated by LLaVA 1.5 13b, while the right panel presents our method's interpretation. While both descriptions recognize the unique setting of a white crater within a desert-like terrain, our method provides a more accurate depiction by identifying the crater as frozen and filled with snow and ice, rather than water. This enhanced understanding highlights our model's ability to handle complex and counterintuitive visual elements, ensuring clarity and correctness in scenarios that defy common expectations.}
   \label{fig:qualitative_results_3}
\end{figure*}

\vspace{2mm}
\noindent
\textbf{Qualitative Examples Across Different Categories.} To better demonstrate the usefulness of our proposed framework, we have included additional qualitative example comparisons such as an animation image in Fig.~\ref{fig:qualitative_results_1}, a documents image in Fig.~\ref{fig:qualitative_results_2}, and a common sense violation image in Fig.~\ref{fig:qualitative_results_3}.

\begin{figure*}[ht]
  \centering
   \includegraphics[width=1\linewidth]{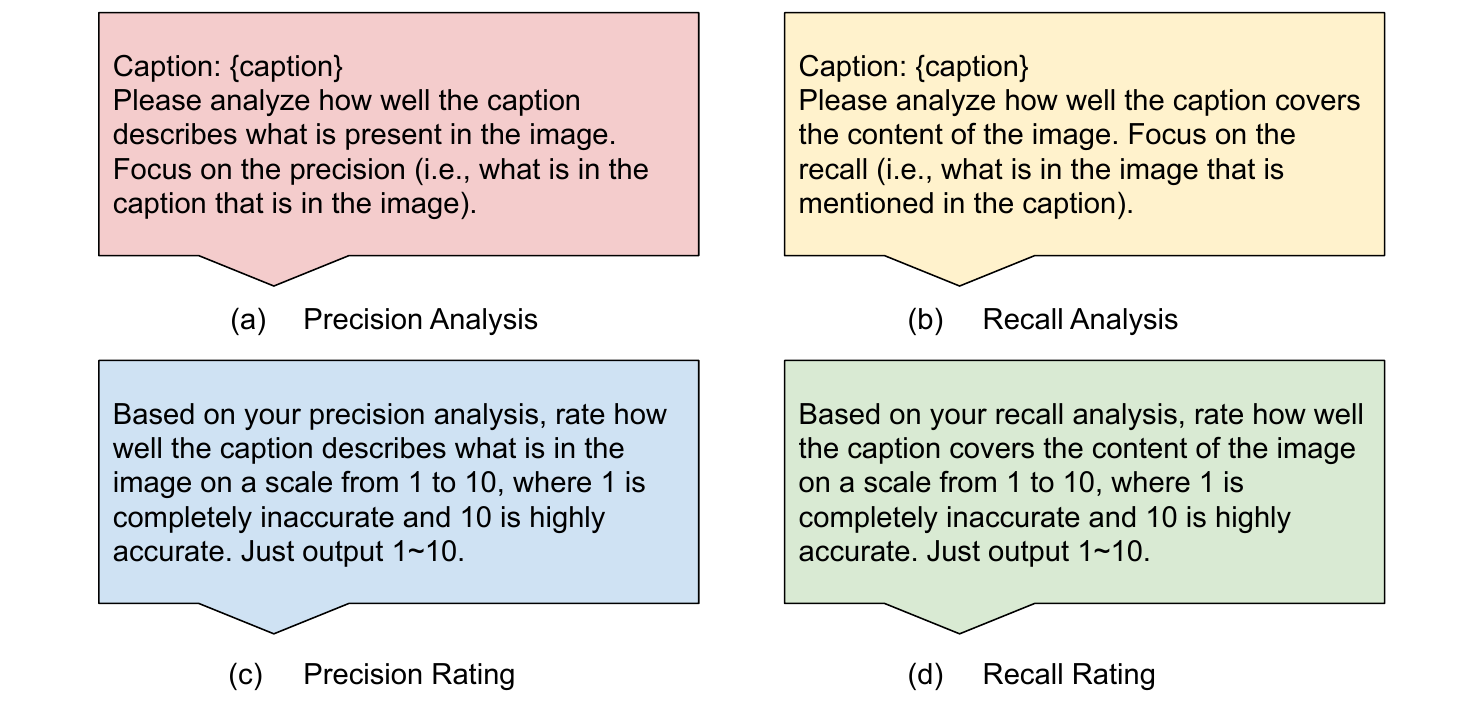}

    \caption{Prompts used for GPT-4o evaluation. We first use (a) and (b) to obtain precision and recall analysis separately, and then, combining those analyses, we use (c) and (d) to calculate the final precision and recall scores accordingly.}
   \label{fig:prompts}
\end{figure*}

\vspace{2mm}
\noindent
\textbf{GPT-4o evaluation.} For the GPT-4o evaluation, each caption was processed by GPT-4o to separately generate precision analysis and recall analysis. The precision analysis was used to compute the precision score, and the recall analysis was used to compute the recall score. The detailed prompts are shown in Fig.~\ref{fig:prompts}.

\begin{figure*}[ht]
  \centering
   \includegraphics[width=1\linewidth]{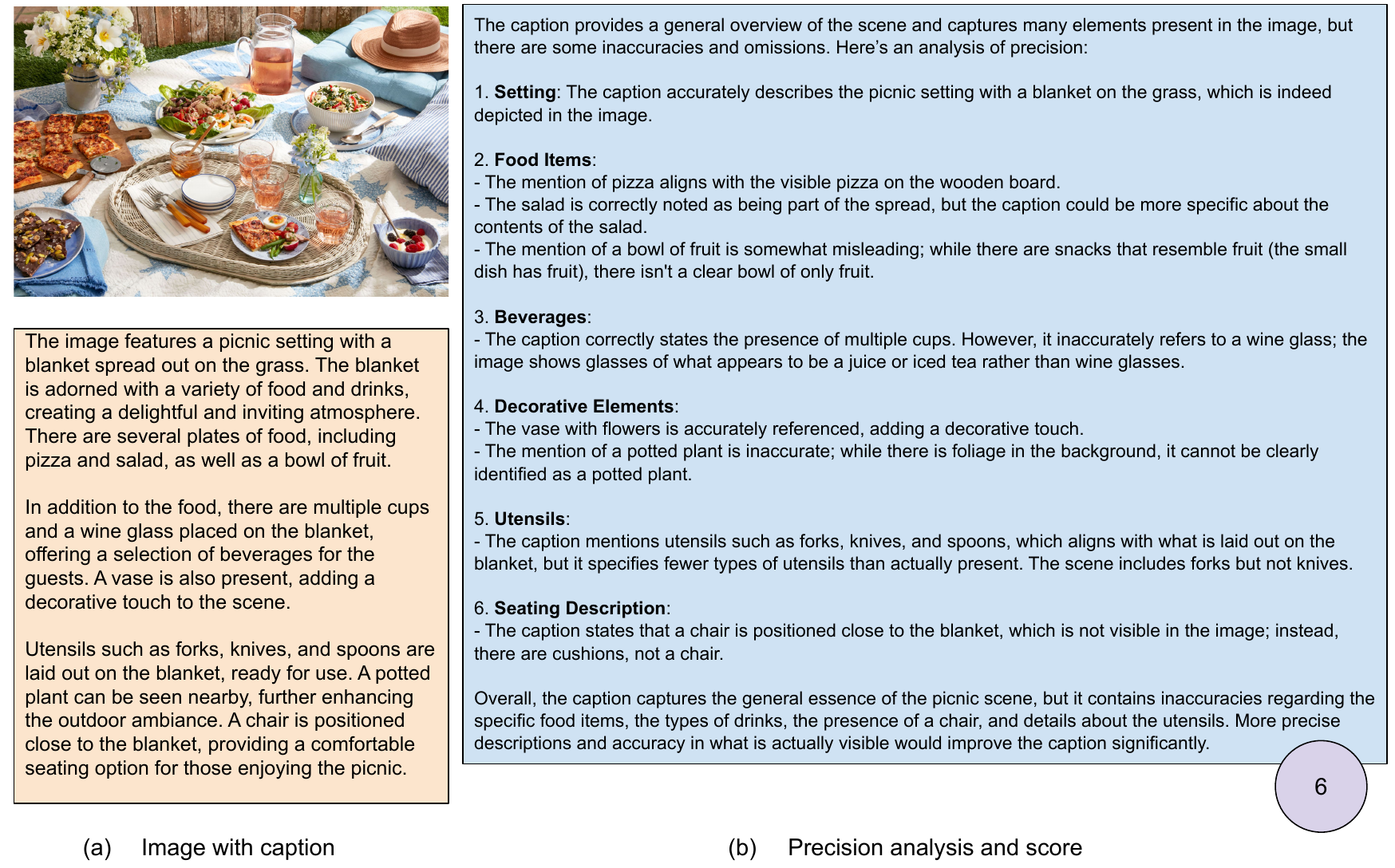}

   \caption{GPT-4o analyzing the precision of a caption given a food image and rating the score.}
   \label{fig:gpt4o_precision}
\end{figure*}

\begin{figure*}[ht]
  \centering
   \includegraphics[width=1\linewidth]{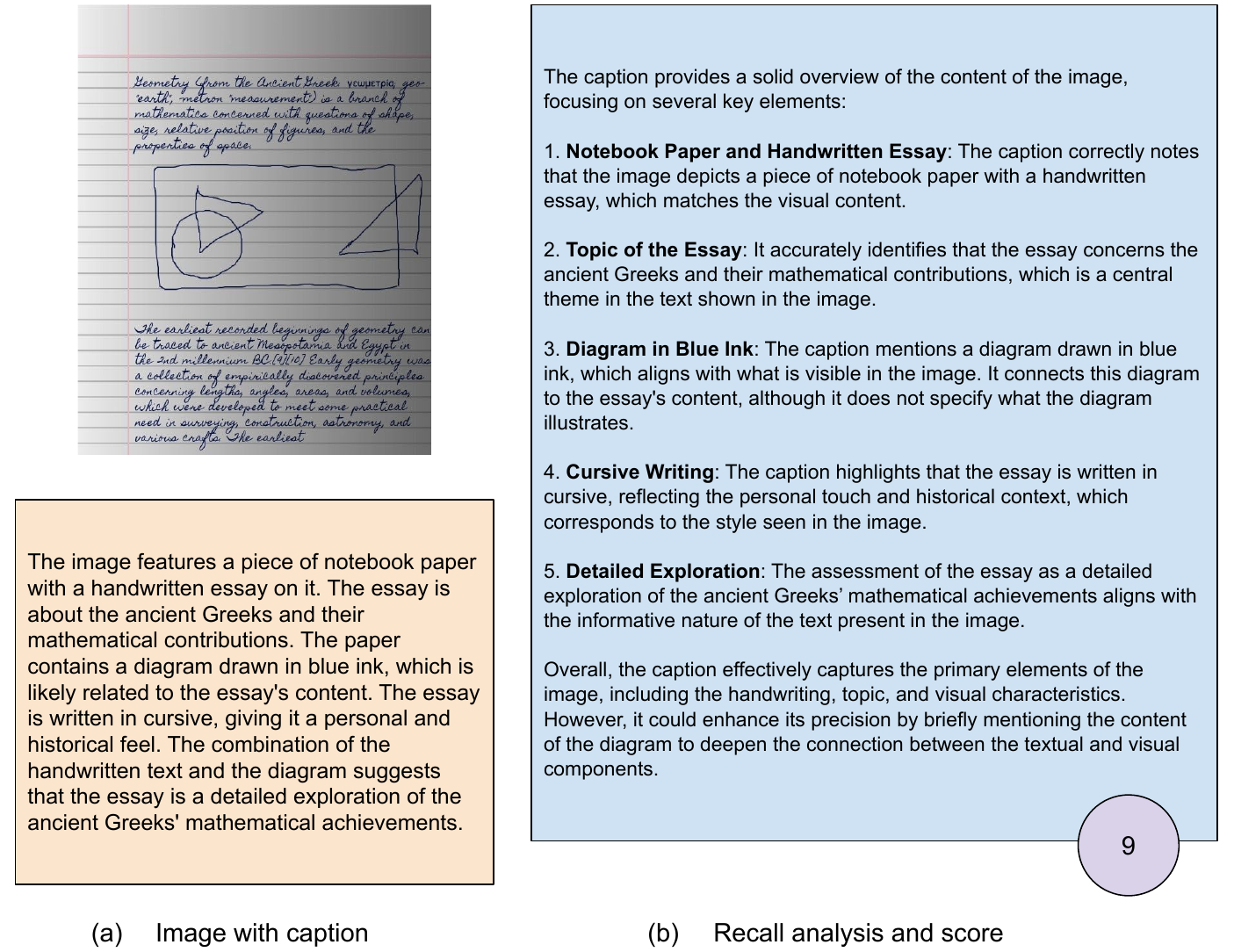}

   \caption{GPT-4o analyzing the recall of a caption given a handwritten image and rating the score.}
   \label{fig:gpt4o_recall}
\end{figure*}

Detailed examples of precision analysis and scores are provided in Fig.~\ref{fig:gpt4o_precision}, and examples of recall analysis and scores are shown in Fig.~\ref{fig:gpt4o_recall}.

\end{document}